\documentclass{article}

\usepackage{microtype}
\usepackage{graphicx}
\usepackage{subcaption}
\usepackage{booktabs} %

\usepackage{hyperref}

\usepackage[preprint]{icml2026}
\hypersetup{
  pdfsubject={Preprint}
}

\usepackage{amsmath}
\usepackage{amssymb}
\usepackage{mathtools}
\usepackage{amsthm}

\usepackage[capitalize,noabbrev]{cleveref}

\theoremstyle{plain}

\theoremstyle{definition}

\theoremstyle{remark}

\usepackage[disable,textsize=tiny]{todonotes}

\icmltitlerunning{On the Optimal Reasoning Length for RL-Trained Language Models}

\begin{document}

\twocolumn[
  \icmltitle{On the Optimal Reasoning Length for RL-Trained Language Models}

  \icmlsetsymbol{equal}{*}

  \begin{icmlauthorlist}
    \icmlauthor{Daisuke Nohara}{isct}
    \icmlauthor{Taishi Nakamura}{isct}
    \icmlauthor{Rio Yokota}{isct}
  \end{icmlauthorlist}

  \icmlaffiliation{isct}{Institute of Science Tokyo, Tokyo, Japan}

  \icmlcorrespondingauthor{Daisuke Nohara}{nohara@rio.scrc.iir.isct.ac.jp}
  \icmlcorrespondingauthor{Rio Yokota}{rioyokota@rio.scrc.iir.isct.ac.jp}

  \icmlkeywords{Language Models, Reinforcement Learning, Reasoning, Test-Time Computation}

  \vskip 0.3in
]

\printAffiliationsAndNotice{}  %

\begin{abstract}
Reinforcement learning substantially improves reasoning in large language models,
but it also tends to lengthen chain-of-thought outputs and increase computational cost.
Although length-control methods have been proposed,
the length--accuracy relationship they induce remains unclear.
We train policies with several length-control methods on multiple base models in a controlled setup and find that,
across both mathematical reasoning and code generation,
accuracy is non-monotonic in output length, peaking at an intermediate value.
Mode accuracy, however, continues to improve with length even in settings where sample accuracy plateaus or declines,
indicating that the non-monotonic length--accuracy relationship is driven by dispersion around an increasingly correct center.
\end{abstract}

\section{Introduction}
Reinforcement learning (RL) applied during post-training has substantially improved the reasoning capabilities of large language models,
enabling extended chain-of-thought (CoT) generation on challenging tasks~\citep{openai2024openaio1card,deepseekai2025deepseekr1incentivizingreasoningcapability,wei2022chain}.
A side effect of this improvement is that RL-trained reasoning models tend to produce increasingly long outputs,
which raises both training and inference costs.
This has motivated a family of length-control methods that penalize verbose generations during
RL~\citep{arora2025training,xiang2025justthinkingefficientreasoning,li2025drpoefficientreasoningdecoupled,shrivastava2025samplethinklessgroup}.
These methods are typically evaluated in terms of efficiency--performance tradeoffs.

A separate line of work has questioned whether longer reasoning is uniformly beneficial.
Test-time interventions that extend the reasoning of a fixed model can degrade
accuracy~\citep{ghosal2025does,su2025underthinkingoverthinkingempiricalstudy},
and the relationship between reasoning length and correctness is often non-monotonic.
It remains unclear whether the same non-monotonicity arises across RL-trained policies
that share the same training setup but differ in their length-control configuration.

To study this, we train policies with several length-control methods at varying penalty strengths,
obtaining a family of policies that span a wide range of average output lengths under an otherwise matched training setup.
Across these policies, accuracy varies non-monotonically with average output length,
rising to a peak at an intermediate length before plateauing or declining.
This pattern holds across three models, four mathematical reasoning benchmarks,
and two code generation benchmarks, and it is not specific to any single length-control method.

A natural question is why longer outputs do not lead to higher accuracy.
Prior work has attributed the test-time version of this phenomenon to dispersion in the answer distribution~\citep{ghosal2025does}.
Within this distributional view, however, a drop in sample accuracy can arise in two different ways.
The center of the distribution may shift away from the correct answer,
or samples may become more dispersed around a center that remains correct.

We find that the second mechanism applies in our setting.
Mode accuracy, defined as the fraction of problems for which the most frequently sampled answer is correct,
continues to improve with average output length even in settings where sample accuracy plateaus or declines.
The distributional center is correct on more problems,
yet individual samples increasingly deviate from it.
We measure this dispersion directly as the average fraction of samples that fall outside the modal group.
This quantity rises with output length in the long-output regime.
The gap between a mode that is correct on more problems and increasingly
dispersed samples explains the non-monotonic length--accuracy relationship.

Our contributions are as follows:
\begin{itemize}
    \item We show that the non-monotonic relationship between output length
      and accuracy, previously documented for test-time interventions on a
      fixed model, also arises across a controlled comparison of RL-trained
      policies spanning multiple models, benchmarks, and length-control
      methods.
  \item We decompose the answer distribution into the correctness of its center
    and the dispersion of samples around that center,
    and show that longer outputs improve the former while increasing the latter.
    This accounts for the otherwise puzzling divergence between mode accuracy and sample accuracy.
\end{itemize}

\section{Related Work}

Recent reasoning models have demonstrated that reinforcement learning can substantially improve reasoning capabilities in LLMs~\citep{openai2024openaio1card,deepseekai2025deepseekr1incentivizingreasoningcapability},
with methods such as GRPO~\citep{GRPO2024Shao} and DAPO~\citep{yu2025dapo}.
While extended chain-of-thought reasoning improves performance on challenging tasks,
RL-trained models tend to produce increasingly long outputs,
raising substantial training and inference costs~\citep{chen2025do}.

This has motivated a family of RL-based length-control methods.
Some apply length-based reward shaping: RLOO-LP~\citep{arora2025training}
and ALP~\citep{xiang2025justthinkingefficientreasoning} penalize longer correct
responses with different normalization strategies.
Others avoid explicit reward shaping: DRPO~\citep{li2025drpoefficientreasoningdecoupled}
uses length-based weighting within the DisCO framework~\citep{li2025disco},
and GFPO~\citep{shrivastava2025samplethinklessgroup} filters samples based on length.
These methods are typically evaluated in terms of efficiency--performance tradeoffs,
asking how much output length can be reduced before accuracy degrades.

A separate line of work questions whether longer reasoning is uniformly beneficial.
\citet{ghosal2025does} analyze a simplified probabilistic model
and show that test-time interventions extending reasoning can degrade accuracy through increased output dispersion.
\citet{su2025underthinkingoverthinkingempiricalstudy} empirically observe a non-monotonic relationship
at the sample level: for a fixed question and model, both excessively short and excessively long generations reduce correctness.
Both studies, however, are conducted at test time on fixed models.
It remains unclear whether the same non-monotonicity arises \emph{across RL-trained policies}
that share the same training setup but differ only in their length-control configuration,
and what mechanism drives the degradation in this setting.

\begin{figure*}[ht]
  \centering
  \includegraphics[
    width=1.0\textwidth,
    keepaspectratio
  ]{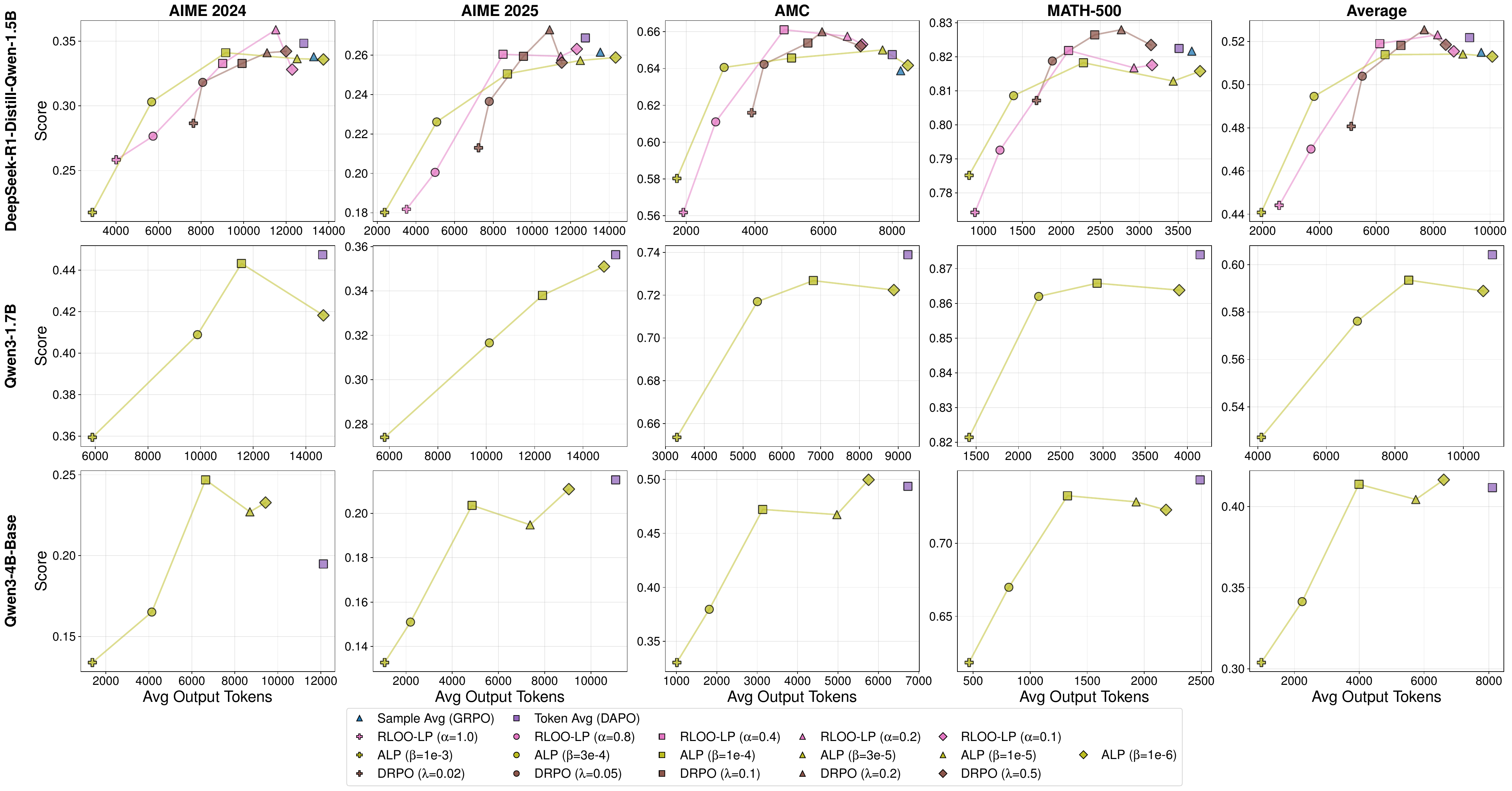}
  \caption{
    \textbf{Accuracy vs.\ average output length on math benchmarks.}
    Each marker corresponds to a policy obtained from RL training with a
    particular length-control method and hyperparameter.
    DeepSeek-R1-Distill-Qwen-1.5B is swept over all five methods;
    Qwen3-1.7B and Qwen3-4B-Base are swept over Token Avg and ALP only.
    Across many model--benchmark pairs, accuracy rises with length, reaches a peak, and then plateaus or declines.
  }
  \label{fig:length_performance_math}
\end{figure*}

\section{Experimental Setup}

\paragraph{Training setup}
We conduct experiments on three models: 
DeepSeek-R1-Distill-Qwen-1.5B~\citep{deepseekai2025deepseekr1incentivizingreasoningcapability},
Qwen3-1.7B, and Qwen3-4B-Base~\citep{yang2025qwen3technicalreport}.
Our training configuration follows DAPO~\citep{yu2025dapo} with two modifications.
First, we disable overlong reward shaping, as length-penalty methods are explored separately in our experiments.
Second, we set the PPO mini-batch size equal to the generation batch size, ensuring fully on-policy updates.

We train on two task domains: mathematical reasoning and code generation.
For mathematical reasoning, we use DAPO-Math-17k and train all three models.
For code generation, we use the DeepCoder-Preview dataset and train DeepSeek-R1-Distill-Qwen-1.5B.

We set the maximum response length to 16K tokens for all models and tasks,
chosen such that fewer than 10\% of rollouts exceed the cap at the start of training.
Within each task setting, all runs are trained for the same number of steps, chosen such that the slowest run takes at least 576 GPU-hours;
some task settings are trained for more steps.
We report results at 480 steps for DeepSeek-R1-Distill-Qwen-1.5B on mathematical reasoning, 180 steps for Qwen3-1.7B on mathematical reasoning,
320 steps for Qwen3-4B-Base on mathematical reasoning, and 360 steps for DeepSeek-R1-Distill-Qwen-1.5B on code generation.
All experiments use FP16 precision~\citep{qi2025defeatingtraininginferencemismatchfp16}.

\paragraph{Evaluation setup}
For mathematical reasoning, we evaluate on four benchmarks: AIME 2024, AIME 2025, AMC, and MATH-500.
For code generation, we evaluate on HumanEval and MBPP+.
Following the evaluation protocol of DeepSeek-R1~\citep{deepseekai2025deepseekr1incentivizingreasoningcapability},
we sample multiple responses per problem.
Further details of the training and evaluation configurations,
along with the rationale for these choices, are provided in Appendix~\ref{app:experimental_details}.

\paragraph{Length-Control Methods}
We first conduct a comprehensive comparison of length-control methods on DeepSeek-R1-Distill-Qwen-1.5B for mathematical reasoning.
As baselines, we consider \textbf{Sample Avg} (the original GRPO objective~\citep{GRPO2024Shao}),
which normalizes the policy gradient loss by response length for each sample,
and \textbf{Token Avg} (the DAPO objective~\citep{yu2025dapo}),
which normalizes by the total token count across the group.
Neither baseline applies an explicit length penalty.
For explicit length penalties, we evaluate three methods.

\textbf{RLOO-LP}~\citep{arora2025training} applies length-based reward shaping to correct responses with per-prompt normalization:
\begin{equation}
\mathcal{R}(x, y) = \mathbf{1}\{y = y^*(x)\} \cdot \left(1 - \alpha \cdot f(\mathrm{len}(y))\right),
\end{equation}
where $f(\mathrm{len}(y)) = \sigma\left((\mathrm{len}(y) - \mathrm{mean}(x))/\mathrm{std}(x)\right)$ normalizes length against the per-prompt distribution of correct response lengths estimated online during training, and $\alpha$ controls the penalty strength.
Larger values of $\alpha$ more strongly favor shorter responses.

\textbf{ALP}~\citep{xiang2025justthinkingefficientreasoning} adjusts the length-penalty strength based on per-prompt accuracy:
\begin{equation}
\mathcal{R}(x, y) = \mathbf{1}\{y = y^*(x)\} - \beta \cdot \mathrm{len}(y) \cdot \max\left(\mathrm{acc}(x), \tfrac{1}{G}\right),
\end{equation}
where $\mathrm{acc}(x)$ is the online per-prompt accuracy, $G$ is the group size, and $\beta$ controls the overall penalty strength. The penalty is stronger on easy (high-accuracy) problems and weaker on difficult ones.

\textbf{DRPO}~\citep{li2025drpoefficientreasoningdecoupled} avoids explicit reward shaping and instead assigns length-based weights to correct responses within the DisCO framework~\citep{li2025disco}:
\begin{equation}
\omega(o) = \exp\left(\frac{1 - |o|/C}{\lambda}\right),
\end{equation}
where $C$ is the maximum response length and $\lambda > 0$ is a regularization parameter; shorter responses receive higher weights,
and smaller values of $\lambda$ more strongly favor shorter responses.
Since these weights are normalized only among correct responses, the learning signal for correct responses remains positive regardless of length.

We also attempted \textbf{GFPO}~\citep{shrivastava2025samplethinklessgroup}, but were unable to reproduce its length-reduction effect (see Appendix~\ref{app:gfpo}).
Since the successfully reproduced methods exhibit qualitatively similar length--performance behavior, we focus on Token Avg and ALP for the remaining model and task settings.
ALP provides a natural continuation of Token Avg, as setting $\beta = 0$ recovers the DAPO objective.
Further details are provided in Appendix~\ref{app:method_details}.

\section{Results}

\subsection{Length--Performance Relationship}
\label{sec:length-perf}

\begin{figure*}[ht]
  \centering
  \includegraphics[
    width=\textwidth,
    keepaspectratio
  ]{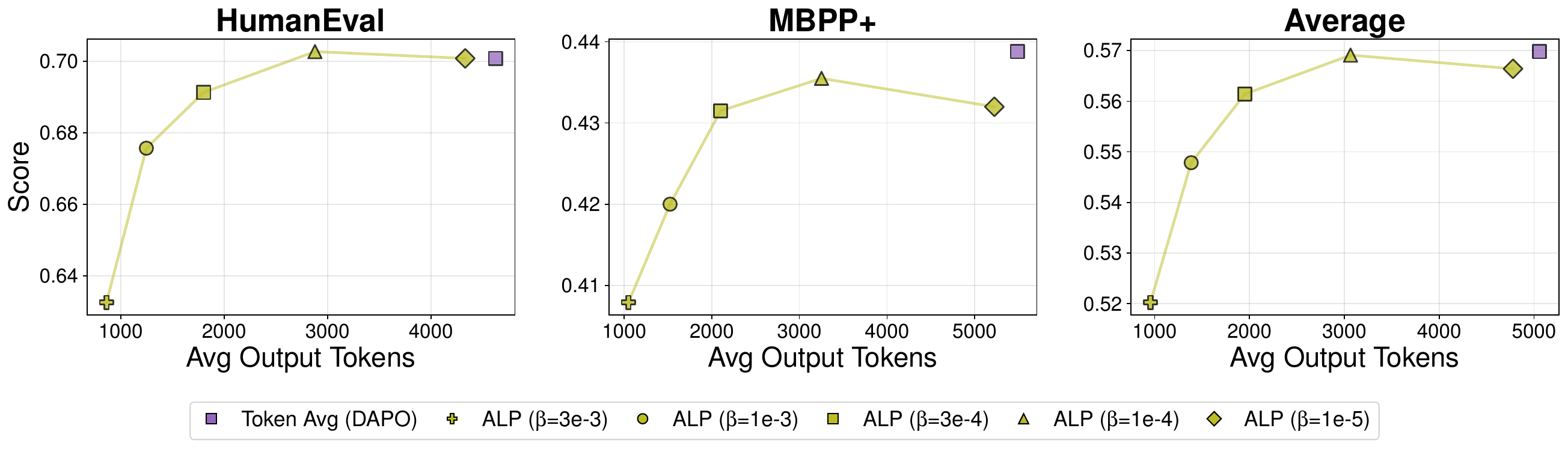}
  \caption{
    \textbf{Accuracy vs.\ average output length on code generation benchmarks.}
    Results for DeepSeek-R1-Distill-Qwen-1.5B trained on DeepCoder-Preview.
    Each marker corresponds to a policy trained with Token Avg (DAPO) or ALP at a particular hyperparameter.
    The same peak-then-decline pattern observed on math benchmarks also appears on code benchmarks.
  }
  \label{fig:length_performance_code}
\end{figure*}

\paragraph{Accuracy is non-monotonic in output length.}
For each length-control method, we sweep its primary hyperparameter
($\alpha$ for RLOO-LP, $\beta$ for ALP, $\lambda$ for DRPO)
to obtain a family of trained policies spanning a range of average output lengths
and plot accuracy as a function of length
in Figures~\ref{fig:length_performance_math} and~\ref{fig:length_performance_code}.
Accuracy is non-monotonic in output length across many of these settings:
it first rises with length, reaches a peak, and then plateaus or declines.
DeepSeek-R1-Distill-Qwen-1.5B shows the pattern across all four math benchmarks:
the best-performing policy occurs at an intermediate length,
and accuracy drops at the longest configurations.
On Qwen3-4B-Base, the pattern is most pronounced on AIME 2024,
where the best ALP configuration reaches 24.7\% accuracy at around 6.6K tokens,
while the unconstrained Token Avg baseline reaches only 19.5\% at over 12K tokens.
Qwen3-1.7B exhibits a milder version within its ALP sweep on every math benchmark except AIME 2025;
on this model, the unconstrained Token Avg baseline matches or slightly exceeds the within-sweep peak.

\paragraph{The pattern is not specific to a single method, model, or domain.}
On DeepSeek-R1-Distill-Qwen-1.5B, where we sweep RLOO-LP, ALP, and DRPO,
the four math benchmarks show a peak followed by a plateau or decline across all three methods,
indicating that the non-monotonicity is not an artifact of any single length-penalty form.
The same pattern appears in the ALP sweeps on Qwen3-1.7B and Qwen3-4B-Base,
so it is not specific to a single model family or scale.
It also extends to code generation on DeepSeek-R1-Distill-Qwen-1.5B
(Figure~\ref{fig:length_performance_code}).
Accuracy peaks and then declines on HumanEval, MBPP+, and their average.

\paragraph{Comparison with prior observations.}
Similar non-monotonic length--accuracy relationships have been reported
in test-time analyses of a single trained model~\citep{ghosal2025does,
su2025underthinkingoverthinkingempiricalstudy}.
In our setting, by contrast, accuracy varies across distinct policies
obtained by RL training with different length-control configurations;
each point on the curve corresponds to a different policy.
Section~\ref{sec:explain} examines the mechanism behind this
non-monotonicity.

\subsection{Explaining the Mode--Sample Divergence}
\label{sec:explain}

\begin{figure*}[ht]
    \centering
    \includegraphics[width=0.95\textwidth]{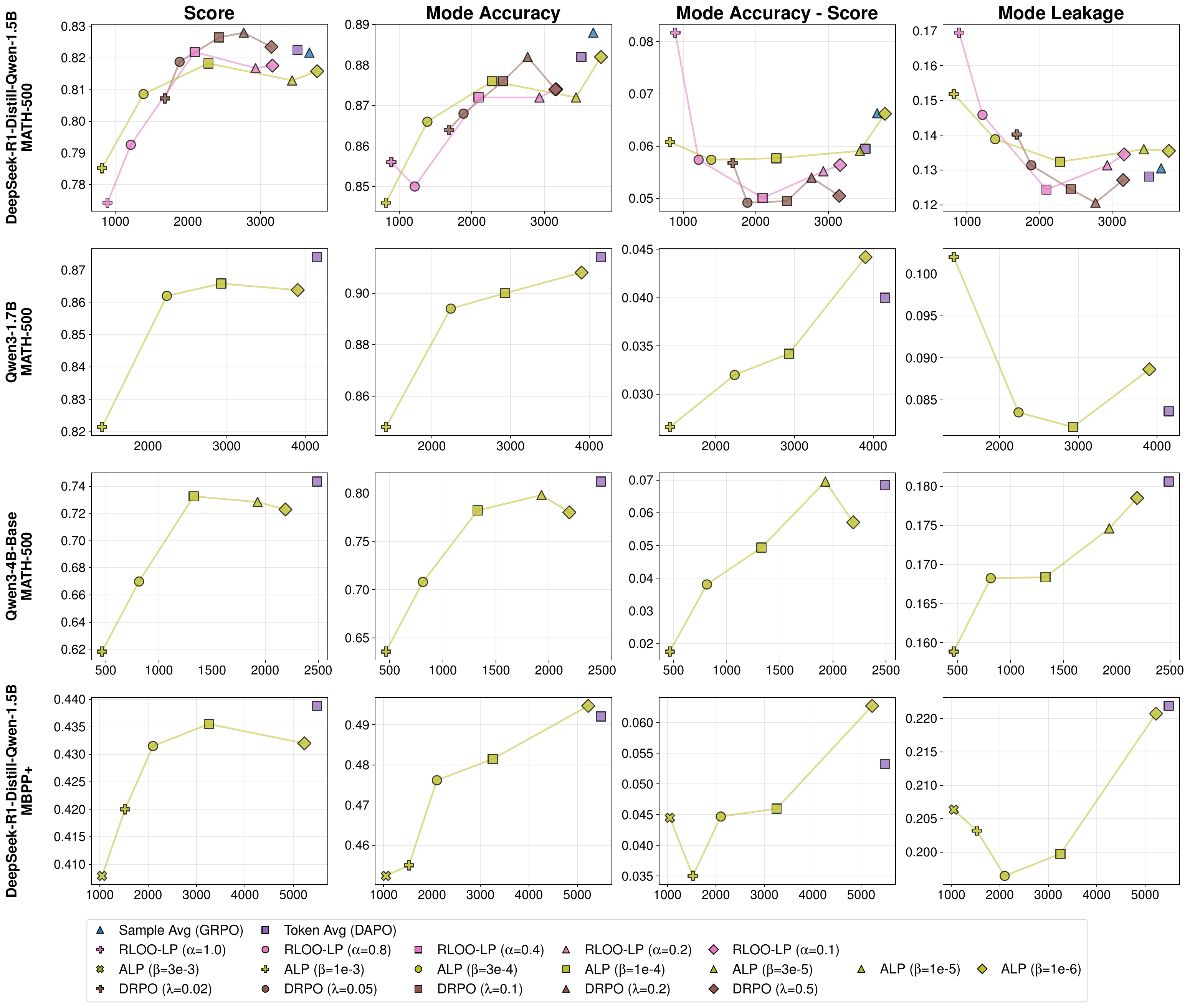}
    \caption{\textbf{Score, mode accuracy, mode accuracy minus score, and
    mode leakage vs.\ average output length} for four representative
    (model, benchmark) pairs. Mode accuracy rises with length in regimes
    where sample accuracy saturates or declines, opening a gap between the two.
    Mode leakage rises in the long-output regime and is U-shaped in three of
    the four settings: at short lengths, samples scatter because the
    distribution has not converged on a dominant answer, while at long
    lengths, samples spread away from an increasingly correct mode.}
    \label{fig:ccl}
\end{figure*}

To better understand the non-monotonic length--accuracy relationship observed in Section~\ref{sec:length-perf},
we analyze it through the variance-based framework of \citet{ghosal2025does}.
They model the distribution over answers $y$ produced by a reasoning model for a question $x$
as a Gaussian $\pi(y \mid x) = \mathcal{N}(\mu_\pi, \sigma_\pi^2)$
centered at $\mu_\pi$ with variance $\sigma_\pi^2$.
The reward function over answers is similarly approximated as $r(y) = \mathcal{N}(\mu_r, \sigma_r^2)$, centered at the correct answer $\mu_r$.
Under this stylized model, the expected reward exhibits a non-monotonic relationship
with policy variance $\sigma_\pi^2$:
small increases in variance improve reward through better coverage of the reward peak,
while excessive variance dilutes probability mass away from it.
They operationalize this dispersion empirically by measuring the entropy of the distribution over extracted answers.

In their analysis, the central tendency $\mu_\pi$ is held fixed and only $\sigma_\pi^2$ is varied.
In our setting, however, different training configurations may shift not only the dispersion but also the center of the answer distribution.
A drop in sample accuracy in this setting could in principle arise from either effect:
the center may shift away from the correct answer, or samples may scatter more widely around the center.
We therefore decompose the behavior into two components:
the correctness of the distributional center and the dispersion of samples around that center.

\paragraph{Metrics.}
For each problem $q$, we draw $N$ samples from the model and group them by answer equivalence:
for math problems, we group by the extracted final answer;
for code, drawing on behavioral clustering
ideas from prior work~\citep{openai2025competitiveprogramminglargereasoning,ravuri2025eliminatinghallucinationinducederrorsllm},
we group solutions by their outputs on a set of test inputs.
We let $\hat{y}_q$ denote the most frequent group (the mode) and $y_q^{\star}$ denote the ground-truth answer.
We write $p_q^{\mathrm{mode}}$ for the fraction of samples in the modal group.
We track two metrics across problems.
\emph{Mode accuracy} is the fraction of problems for which $\hat{y}_q = y_q^{\star}$,
capturing whether the center of the answer distribution is correct.
\emph{Mode leakage} is the average fraction of samples that fall outside the modal group:
\[
    \mathrm{ModeLeakage}
    = \frac{1}{|Q|} \sum_{q \in Q}
      \left(1 - p_q^{\mathrm{mode}}\right).
\]
Intuitively, mode leakage measures how dispersed samples are around the center of the answer distribution,
in parallel with the entropy-based dispersion measure of \citet{ghosal2025does}.

\paragraph{Mode accuracy improves with length while sample accuracy plateaus or declines.}
Figure~\ref{fig:ccl} plots score, mode accuracy, their gap, and mode leakage against average output length
on MATH-500 for all three models and on MBPP+ for DeepSeek-R1-Distill-Qwen-1.5B;
corresponding results on AIME 2024, AIME 2025, AMC, and HumanEval are reported in Appendix~\ref{app:bias-variance-full}.
Across these settings, mode accuracy rises with output length more steadily than sample accuracy.
In the long-output regime, where sample accuracy saturates or declines,
mode accuracy continues to rise, opening a gap between the two.
The center of the answer distribution becomes more correct with length,
but individual samples do not track this improvement.

\paragraph{Mode leakage accounts for the divergence.}
On Qwen3-4B-Base, mode leakage rises monotonically with output length,
mirroring the saturation of sample accuracy as mode accuracy continues to climb.
On DeepSeek-R1-Distill-Qwen-1.5B (both MATH-500 and MBPP+) and on Qwen3-1.7B,
mode leakage is U-shaped: it is minimal near the configurations with the highest sample accuracy
and elevated at both shorter and longer lengths.
The two arms of the U correspond to qualitatively distinct mechanisms.
At short lengths, mode accuracy is also low: the answer distribution has not yet converged on a dominant answer,
and the elevated mode leakage reflects this lack of concentration.
At long lengths, mode accuracy is at its highest,
yet mode leakage rises again as samples spread away from an increasingly correct mode.

These results refine the variance-based account of \citet{ghosal2025does} for our setting:
longer outputs not only increase the spread of the answer distribution
but also shift its center toward the correct answer.
In the long-output regime, both mode accuracy and mode leakage rise,
and sample accuracy reflects the net effect of these opposing forces:
gains from a more correct center eventually fail to offset losses from samples spreading away from it.

\section{Conclusion}
We studied how output length affects the performance of RL-trained reasoning models
by training policies under several length-control methods
at varying penalty strengths and comparing them as a function of average output length.
Across three models and two task domains,
we found that sample accuracy is non-monotonic in output length,
while mode accuracy continues to rise into the long-output regime.
By decomposing the answer distribution into the correctness of its center
and the dispersion of samples around that center,
we showed that longer outputs improve the former while increasing the latter,
and that this dispersion around an increasingly correct center accounts for the non-monotonic length--accuracy relationship.
A natural direction for future work is to develop length-control methods
that automatically identify the optimal length regime without manual hyperparameter search,
since the optimal penalty strength varies across methods and models.

\section*{Acknowledgements}

This work was supported by JST K Program Japan Grant Number JPMJKP24C3. This work used computational resources TSUBAME4.0 supercomputer provided by Institute of Science Tokyo through the HPCI System Research Project (Project ID: hp260015).
We thank Masaki Kawamura for helpful discussions on the experiments and feedback on the manuscript.

\section*{Impact Statement}

This paper presents work whose goal is to advance the field of Machine
Learning. There are many potential societal consequences of our work, none of
which we feel must be specifically highlighted here.

\bibliography{example_paper}
\bibliographystyle{icml2026}

\newpage
\appendix
\onecolumn

\section{Experimental Details}
\label{app:experimental_details}

\subsection{Training}
\label{app:training_details}
We train three models on two task domains: DeepSeek-R1-Distill-Qwen-1.5B, Qwen3-1.7B, and Qwen3-4B-Base on mathematical reasoning, and DeepSeek-R1-Distill-Qwen-1.5B on code generation.
Mathematical reasoning experiments use the DAPO-Math-17K dataset, and code generation experiments use the DeepCoder-Preview dataset.
All experiments are conducted using the verl framework.

Our configuration adopts GRPO with the KL penalty removed ($\beta=0$), computing group-relative advantages by normalizing rewards across sampled responses per prompt (subtracting mean and dividing by standard deviation).
DAPO-style dynamic sampling is enabled to filter out prompts with zero gradient signal.
For optimization, we use AdamW with a learning rate of $1 \times 10^{-6}$, weight decay of 0.1, and 10 warmup steps.
We use a prompt batch size of 64, sampling 16 responses per prompt, with generation temperature 1.0.
The rationale for setting the PPO mini-batch size equal to the generation batch size is discussed in Appendix~\ref{app:batch_size}.

The maximum response length is set to 16K tokens for all models and tasks, chosen such that fewer than 10\% of rollouts exceed the cap at the start of training.
Within each task setting, all runs are trained for the same number of steps, chosen such that the slowest run takes at least 576 GPU-hours under an 8-GPU setup (72 hours of wall-clock time); some task settings are trained for more steps.
We report results at 480 steps for DeepSeek-R1-Distill-Qwen-1.5B on mathematical reasoning, 180 steps for Qwen3-1.7B on mathematical reasoning, 320 steps for Qwen3-4B-Base on mathematical reasoning, and 360 steps for DeepSeek-R1-Distill-Qwen-1.5B on code generation.
We use FSDP for distributed training and vLLM for rollout generation.

\paragraph{Preliminary experiments on Qwen3-1.7B-Base.}
We additionally use Qwen3-1.7B-Base for preliminary experiments on training dynamics (Appendix~\ref{app:batch_size}) and for our GFPO reproduction attempt (Appendix~\ref{app:gfpo}).
For these experiments, the maximum response length was set to 8K tokens, and training was conducted in BF16 with truncated importance sampling (TIS)~\citep{yao2025offpolicy}.

\subsection{Evaluation}
\label{app:eval_details}
We use the same prompt template as the training dataset for mathematical reasoning to ensure consistency between training and evaluation.
For code generation, we use a slightly different prompt template at evaluation, while keeping the answer extraction procedure consistent with training.
For mathematical reasoning, we sample 64 responses per problem for AIME 2024, AIME 2025, and AMC, which contain fewer problems, and 16 responses per problem for MATH-500.
For code generation, we sample 16 responses per problem for HumanEval, MBPP, and MBPP+.
MBPP results are reported in Appendix~\ref{app:bias-variance-full} as a supplementary check.
Following DeepSeek-R1~\citep{deepseekai2025deepseekr1incentivizingreasoningcapability}, we set the temperature to 0.6 and top-$p$ to 0.95, with a context size of 32K tokens.
Evaluation is conducted using the EvalScope framework with vLLM for efficient inference.

\subsection{Batch Size Configuration}
\label{app:batch_size}
This section justifies our choice in the main experiments to set the PPO mini-batch size equal to the generation prompt batch size, eliminating off-policy updates during training.
The supporting experiments here were conducted on Qwen3-1.7B-Base in BF16 with truncated importance sampling (TIS) as part of our preliminary studies.

\begin{figure*}[t]
  \centering
  \includegraphics[width=\linewidth]{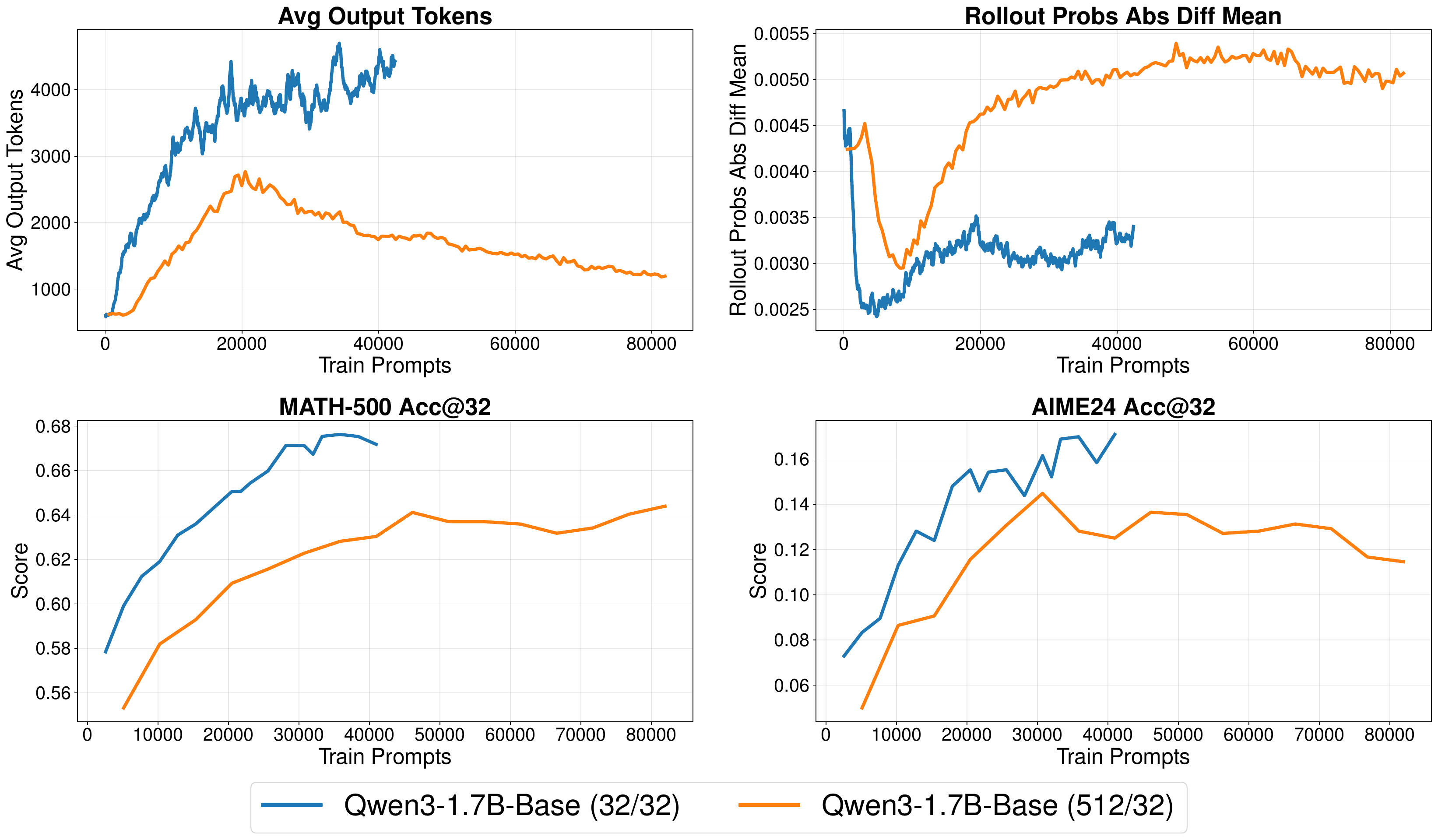}
  \caption{
    \textbf{Effect of batch size configuration on training dynamics for Qwen3-1.7B-Base (BF16 with TIS).}
    Top left: response length during training. Top right: absolute difference between rollout and training token probabilities.
    Bottom: validation scores on MATH-500 and AIME 2024.
    The 512/32 setting (generation batch size 512, mini-batch size 32) leads to decreasing response length and validation performance, while the 64/64 setting maintains stable training.
  }
  \label{fig:batch_size}
\end{figure*}

Initially, we followed DAPO's default configuration with a generation prompt batch size of 512 and a PPO mini-batch size of 32.
Under this 512/32 setting, we observed that the response length began to decrease during training, accompanied by a decline in validation scores on MATH-500 and AIME 2024 (Figure~\ref{fig:batch_size}).
To address this issue, we set the generation prompt batch size equal to the PPO mini-batch size, thereby eliminating off-policy updates.
With this configuration, the response length consistently increased throughout training, and the validation performance remained stable.

We also measured the absolute difference between token probabilities computed by the rollout engine and the training engine.
As shown in Figure~\ref{fig:batch_size} (top right), this difference increased substantially under the 512/32 setting, suggesting that the mismatch between rollout and training policies contributes to the instability.
It is known that numerical differences between the rollout engine and training engine can cause training instability; our results suggest that the probability difference grows substantially when multiple gradient updates are performed per rollout batch.

For the main experiments, we chose a batch size of 64 rather than 32 to maximize rollout efficiency while maintaining training stability, and we confirmed empirically that this setting does not exhibit the instability observed with the 512/32 configuration.

\subsection{Precision and Truncated Importance Sampling}
\label{app:precision}
All main experiments are conducted in FP16 without truncated importance sampling (TIS).

\begin{figure*}[t]
  \centering
  \includegraphics[width=\linewidth]{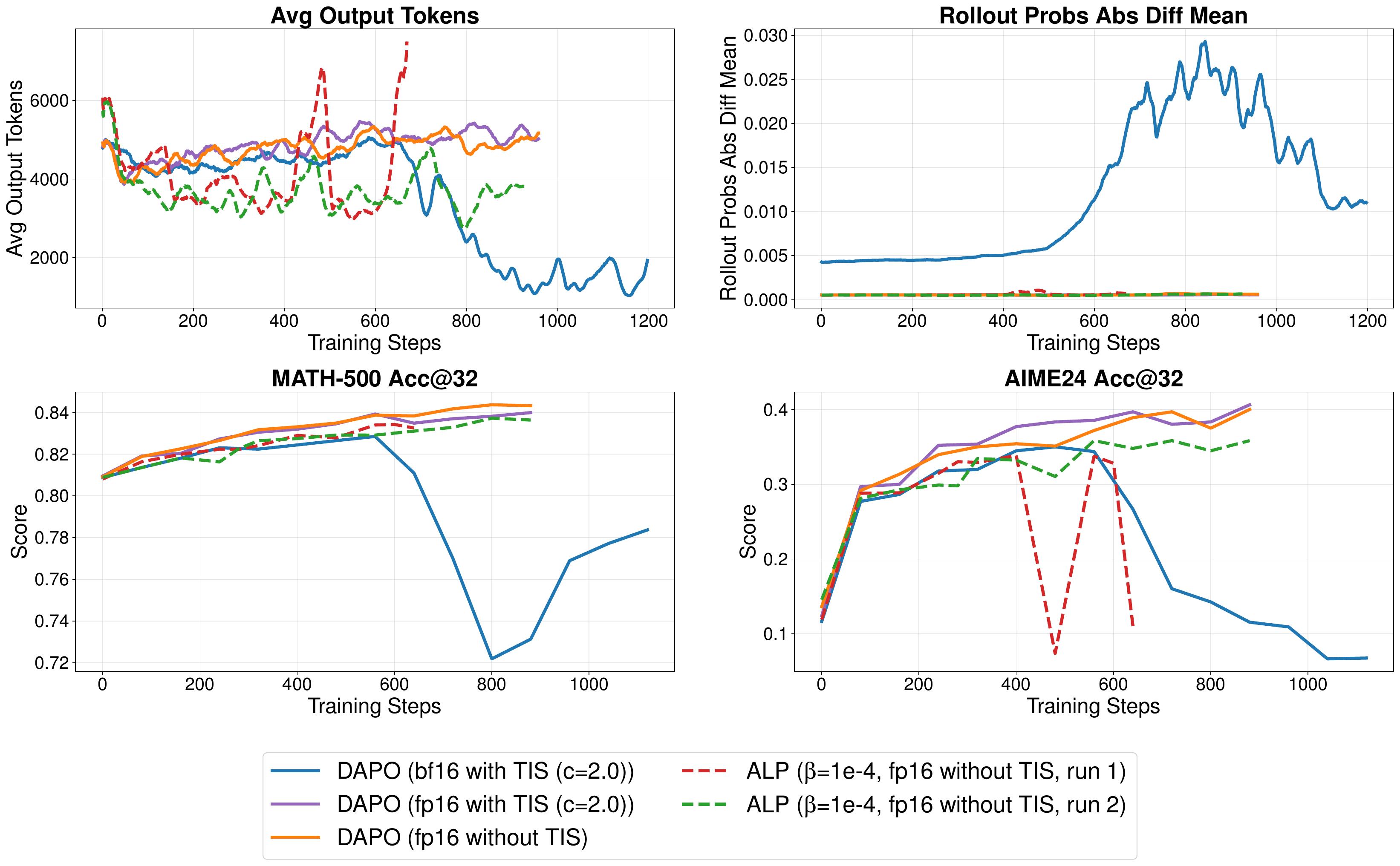}
  \caption{
    \textbf{Effect of precision and TIS on training dynamics for DeepSeek-R1-Distill-Qwen-1.5B.}
    Comparison of BF16 with TIS, FP16 with TIS, FP16 without TIS, and ALP ($\beta=1\mathrm{e}{-4}$) in FP16 without TIS.
    Note that the maximum response length differs between the ALP experiment and the precision ablation experiments, so response lengths should not be directly compared across these settings.
  }
  \label{fig:precision}
\end{figure*}

For DeepSeek-R1-Distill-Qwen-1.5B, we initially attempted to train in BF16 with TIS, but encountered training instability despite using TIS: training diverged in both of two independent runs.
Following \citet{qi2025defeatingtraininginferencemismatchfp16}, we switched to FP16, which substantially reduced the probability difference between the rollout and training engines (Figure~\ref{fig:precision}) and enabled stable training.
We conducted an ablation comparing FP16 with and without TIS, and found no noticeable difference in the probability gap.
For simplicity, we chose to train without TIS in FP16.
For Qwen3-1.7B and Qwen3-4B-Base, FP16 without TIS likewise produced stable training, so we adopted the same configuration across all main experiments.

Even with FP16, training instability can still occur in rare cases.
In our length-penalty experiments, ALP with $\beta=1\mathrm{e}{-4}$ exhibited instability: the probability difference temporarily spiked to a large value, causing training to fail.
Although this spike is smaller than the BF16--FP16 gap, it was sufficient to destabilize training.
We restarted training from scratch with the same configuration, and the second run completed successfully, which we report as our result.
The instability with ALP does not occur consistently across runs.

\section{Details of Length-Control Methods}
\label{app:method_details}

\subsection{RLOO-LP}

RLOO-LP ~\citep{arora2025training} applies length-based reward shaping to correct responses with per-prompt normalization with RLOO (REINFORCE Leave-One-Out) advantage estimation.

RLOO samples $n$ responses $\{y_1, \ldots, y_n\}$ for a prompt $x$ and estimates the advantage for each response $y_i$ as:
\begin{equation}
\hat{A}(y_i, x) = \mathcal{R}(y_i, x) - \frac{1}{n-1} \sum_{j \neq i} \mathcal{R}(y_j, x)
\end{equation}

~\cite{arora2025training} define a reward function that applies a length penalty to correct responses:
\begin{equation}
\mathcal{R}(x, y) = \mathbf{1}\{y = y^*(x)\} \cdot \left(1 - \alpha \cdot f(\mathrm{len}(y))\right)
\end{equation}
where $\alpha$ is a hyperparameter controlling the penalty strength;
in our experiments, we sweep values up to $\alpha=1.0$.
The normalization function $f(\cdot)$ is defined as:
\begin{equation}
f(\mathrm{len}(y)) = \sigma\left(\frac{\mathrm{len}(y) - \mathrm{mean}(x)}{\mathrm{std}(x)}\right)
\end{equation}
where $\mathrm{mean}(x)$ and $\mathrm{std}(x)$ are the mean and standard deviation of correct response lengths for prompt $x$, estimated online during training. This per-prompt normalization prevents excessively penalizing longer reasoning required for difficult problems. When $\alpha = 0$, the reward function reduces to the standard RLVR setting.
Larger values of $\alpha$ more strongly favor shorter responses.

\subsection{ALP}

Adaptive Length Penalty (ALP)~\cite{xiang2025justthinkingefficientreasoning} adjusts the length penalty strength based on per-prompt accuracy. The reward function is defined as:
\begin{equation}
\mathcal{R}(x, y) = \mathbf{1}\{y = y^*(x)\} - \beta \cdot \mathrm{len}(y) \cdot \max\left(\mathrm{acc}(x), \frac{1}{G}\right)
\end{equation}
where $\mathrm{acc}(x)$ is the online accuracy for prompt $x$ estimated during training, $G$ is the group size (number of samples per prompt), and $\beta$ is a hyperparameter controlling the overall penalty strength. For high-accuracy (easy) problems, the length penalty is stronger, while for low-accuracy (difficult) problems, the penalty is weaker. The $\max(\mathrm{acc}(x), 1/G)$ term ensures a minimum penalty is applied even when accuracy is zero.

\subsection{DRPO}

DRPO (Decoupled Reward Policy Optimization)~\citep{li2025drpoefficientreasoningdecoupled} assigns length-based weights to correct responses within the DisCO framework~\citep{li2025disco}, ensuring that learning signals remain positive regardless of response length.

\paragraph{DisCO Objective.}
DisCO defines a score function as the average log-likelihood of a response:
\begin{equation}
\label{eq:disco_score}
s_\theta(o, q) = \frac{1}{|o|} \sum_{t=1}^{|o|} \log \pi_\theta(o_t | q, o_{<t})
\end{equation}
The DisCO objective is:
\begin{equation}
\max_\theta \mathbb{E}_q \left[ \mathbb{E}_{o \in \mathcal{C}_q} s_\theta(o, q) - \tau \log \sum_{o' \in \mathcal{W}_q} \exp\left(\frac{s_\theta(o', q)}{\tau}\right) \right]
\end{equation}
where $\mathcal{C}_q$ is the set of correct responses for prompt $q$, $\mathcal{W}_q$ is the set of incorrect responses, and $\tau$ is a temperature parameter.

\paragraph{DRPO Objective.}
DRPO introduces length-based weights for correct responses:
\begin{equation}
\omega(o) = \exp\left(\frac{1 - |o|/C}{\lambda}\right)
\end{equation}
where $C$ is the maximum response length and $\lambda > 0$ is a regularization parameter. Shorter responses receive higher weights. The DRPO objective is:
\begin{equation}
\max_\theta \mathbb{E}_q \left[ \frac{\sum_{o \in \mathcal{C}_q} \omega(o) \cdot s_\theta(o, q)}{\sum_{o \in \mathcal{C}_q} \omega(o)} - \tau \log \sum_{o' \in \mathcal{W}_q} \exp\left(\frac{s_\theta(o', q)}{\tau}\right) \right]
\end{equation}
Since the weights are normalized only among correct responses, the learning signal for correct responses remains positive regardless of length. Smaller $\lambda$ more strongly favors shorter responses, while $\lambda \to \infty$ recovers the DisCO objective.

\subsection{GFPO}

GFPO (Group Filtered Policy Optimization)~\citep{shrivastava2025samplethinklessgroup} mitigates length inflation by sampling a larger group of responses for each problem and filtering the sampled responses before applying policy updates.

For each prompt $x$, GFPO samples a group of $G$ responses $\{y_i\}_{i=1}^{G}$ from the current policy. Each response is scored using a filtering metric $m(x, y_i)$, such as response length or token efficiency. A filtered subset $\mathcal{F}_x$ is then constructed by retaining the top-$k$ responses according to this metric:
\[
\mathcal{F}_x
=
\operatorname{TopK}_{y_i \in \{y_1,\ldots,y_G\}}
m(x, y_i).
\]
Policy updates are computed only from responses in $\mathcal{F}_x$, while responses outside the filtered set are masked out by assigning them zero advantage. The resulting objective follows the GRPO framework, but with advantage estimation restricted to the filtered subset. In the length-based variant, the filter favors shorter responses; in the token-efficiency variant, responses are ranked by reward per token, thereby encouraging concise responses that still achieve high reward.

\subsubsection{GFPO Reproduction Attempt}
\label{app:gfpo}

We attempted to evaluate the length-based filtering variant of GFPO~\citep{shrivastava2025samplethinklessgroup} in our experimental setup. Specifically, we implemented the simplest configuration: sampling $G=16$ responses per prompt and selecting only the $k=8$ shortest outputs for training. We applied this method to both Qwen3-1.7B-Base and DeepSeek-R1-Distill-Qwen-1.5B. However, we were unable to reproduce the length-reduction effect reported in the original work.

\begin{figure}[ht]
  \centering
  \includegraphics[width=\linewidth]{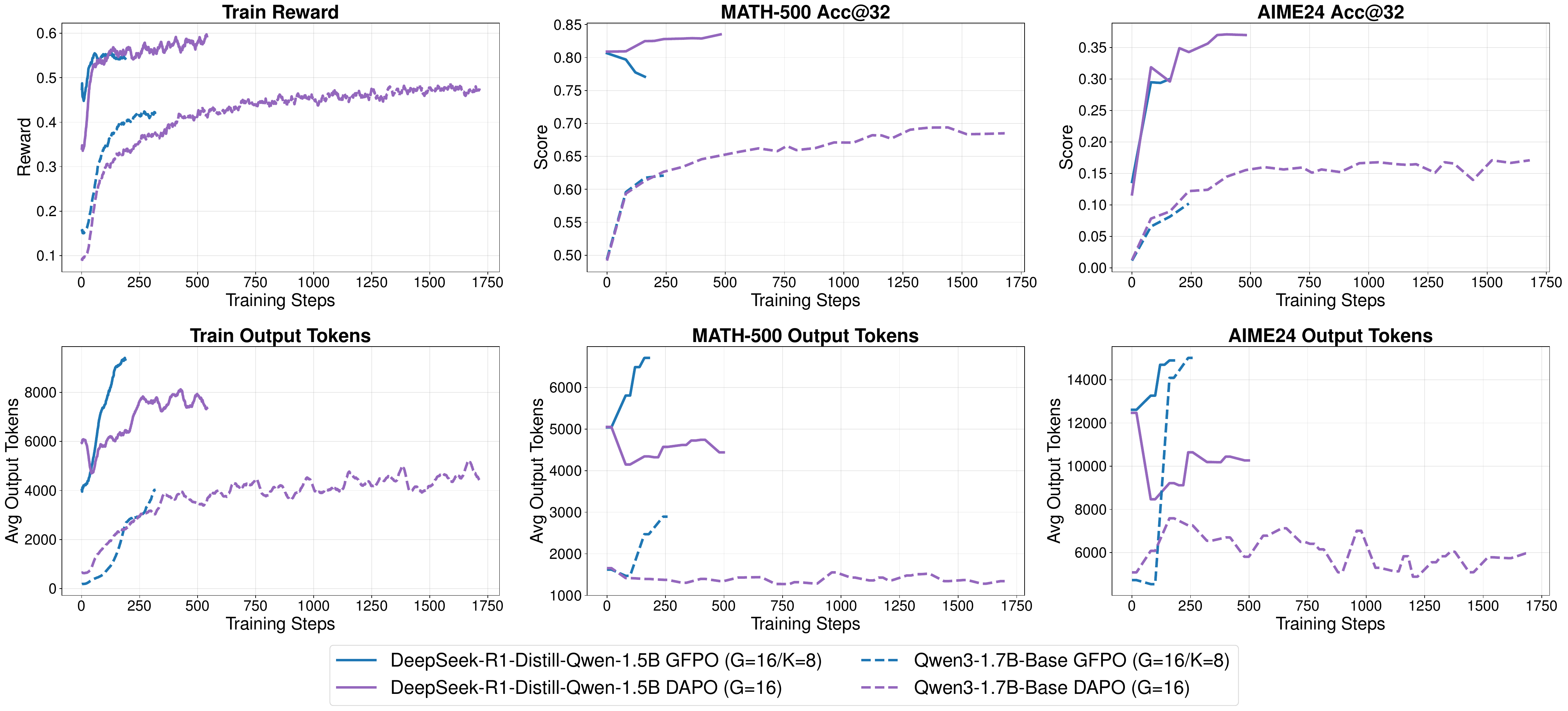}
  \caption{
    \textbf{GFPO reproduction attempt on two base models.}
    Average output length during training for GFPO ($G=16$, $k=8$) compared to the DAPO baseline. Both Qwen3-1.7B-Base and DeepSeek-R1-Distill-Qwen-1.5B show increasing output length under GFPO, particularly in later training stages.
  }
  \label{fig:gfpo}
\end{figure}

As shown in Figure~\ref{fig:gfpo}, the average output length on the training dataset increased throughout training for both models, particularly in the later stages, compared to the DAPO baseline without any length penalty. This effect was even more pronounced on the validation dataset.

We note that GFPO appears to produce shorter outputs than DAPO in early training in Figure~\ref{fig:gfpo}. However, this is an artifact of our logging implementation, which computes the average length \emph{after} filtering for the shortest outputs rather than across all sampled responses.

We hypothesize that GFPO's filtering mechanism, which excludes longer outputs from training,
also prevents the model from using long incorrect responses as a learning signal.
In settings where explicitly penalizing verbose failures is necessary to control output length, GFPO may not be effective.

Due to this inability to reproduce the expected behavior, we exclude GFPO from our main experimental comparisons.

\section{GRPO and DAPO Objectives}
\label{app:grpo_dapo}

Group Relative Policy Optimization (GRPO)~\citep{GRPO2024Shao} eliminates the
need for a separate value model by computing group-relative advantages from
multiple sampled responses per prompt.
GRPO normalizes the loss by response length for each sample:
\begin{equation}
  J_{\text{GRPO}}(\theta) = \mathbb{E} \left[
    \frac{1}{G} \sum_{i=1}^{G} \frac{1}{|y_i|} \sum_{t=1}^{|y_i|}
    \min \left( \rho_{i,t} \hat{A}_{i,t},\;
    \text{clip}(\rho_{i,t}, 1-\epsilon, 1+\epsilon) \hat{A}_{i,t} \right)
  \right]
\end{equation}
where $G$ is the number of responses per prompt and $|y_i|$ is the length of
response $i$.

DAPO~\citep{yu2025dapo} modifies this by normalizing by the total token count:
\begin{equation}
  J_{\text{DAPO}}(\theta) = \mathbb{E} \left[
    \frac{1}{\sum_{i=1}^{G} |y_i|} \sum_{i=1}^{G} \sum_{t=1}^{|y_i|}
    \min \left( \rho_{i,t} \hat{A}_{i,t},\;
    \text{clip}(\rho_{i,t}, 1-\epsilon, 1+\epsilon) \hat{A}_{i,t} \right)
  \right]
\end{equation}
These normalization schemes can affect training stability depending on the
variance of output lengths, as analyzed in the next section.

\subsection{Analysis of Output Length Variance}
\label{app:variance}

We analyze why Sample Avg (GRPO) and DRPO exhibit unstable training on Qwen3-1.7B-Base compared to DeepSeek-R1-Distill-Qwen-1.5B.

\begin{figure*}[t]
  \centering
  \includegraphics[width=\linewidth]{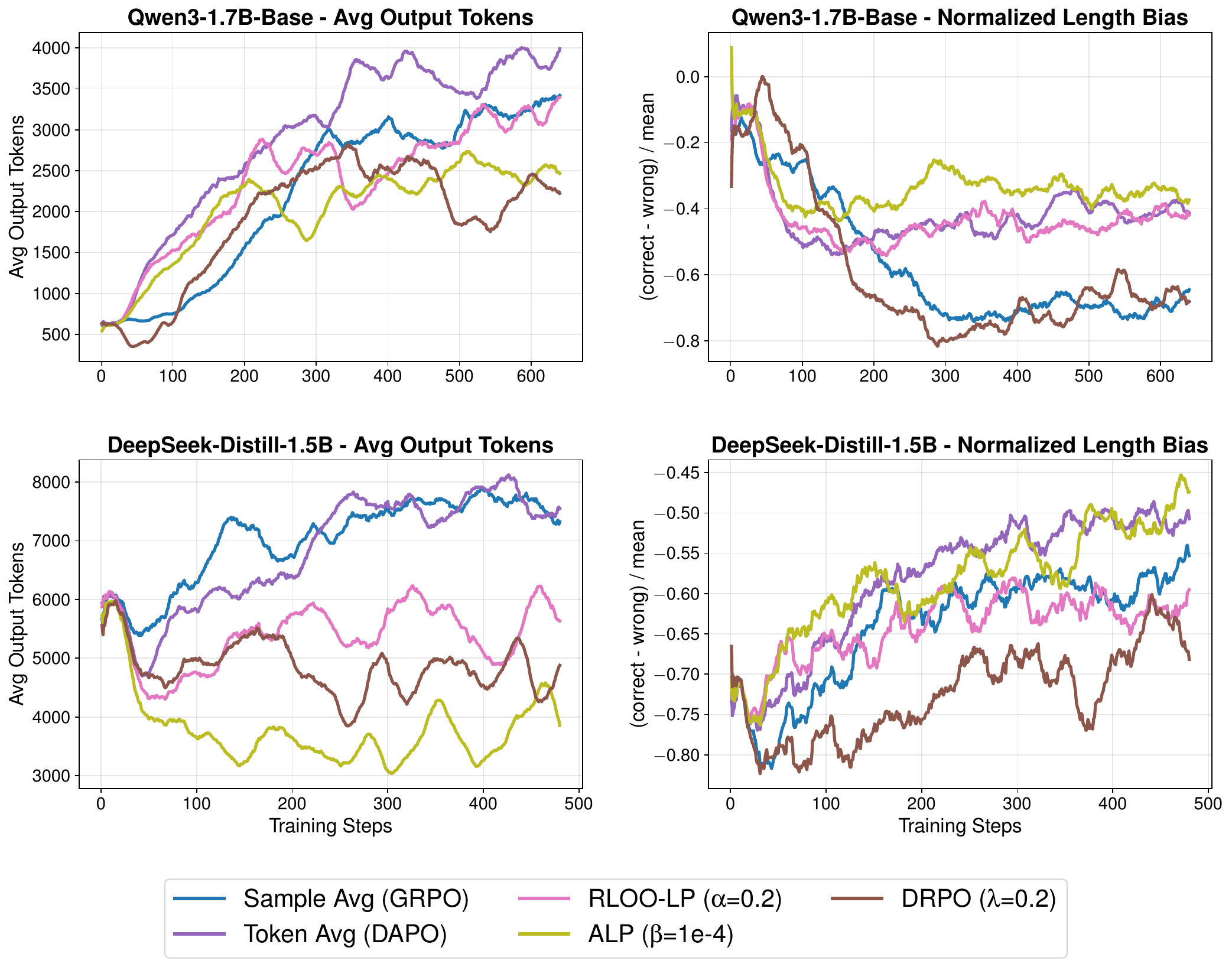}
  \caption{
    \textbf{Evolution of output length during training} (20-step moving average).
    Left: average output length. Right: length bias (normalized difference between correct and incorrect response lengths).
    On Qwen3-1.7B-Base, Sample Avg (GRPO) and DRPO exhibit large negative length bias, indicating that incorrect responses tend to be longer than correct ones.
  }
  \label{fig:length_curve}
\end{figure*}

Figure~\ref{fig:length_curve} shows the evolution of average output length and length bias during training. Length bias is defined as the difference between mean lengths of correct and incorrect responses, normalized by the overall mean length. A negative length bias indicates that incorrect responses tend to be longer than correct ones. On Qwen3-1.7B-Base, Sample Avg (GRPO) and DRPO exhibit large negative length bias, suggesting insufficient suppression of long incorrect responses.

Table~\ref{tab:length_cv} shows the coefficient of variation (CV) of output lengths during the first 10 training steps. We report the overall CV computed from all samples, the within-prompt CV averaged across prompts, and the between-prompt CV computed from per-prompt mean lengths.

\begin{table}[ht]
\centering
\caption{Coefficient of variation of output lengths (Token Avg, first 10 steps).}
\label{tab:length_cv}
\begin{tabular}{lccc}
\toprule
Model & Overall & Within-prompt & Between-prompt \\
\midrule
Qwen3-1.7B-Base & 1.53 & 1.10 & 0.45 \\
DeepSeek-R1-Distill-1.5B & 0.68 & 0.45 & 0.50 \\
\bottomrule
\end{tabular}
\end{table}

Qwen3-1.7B-Base exhibits substantially higher within-prompt CV compared to DeepSeek-R1-Distill-Qwen-1.5B. In Sample Avg, each sample is weighted by the inverse of its output length $1/|o|$; when within-prompt length variance is high, the weight differences become pronounced, leading to unstable gradient updates. DRPO exhibits similar instability because its underlying framework, DisCO, uses the same per-sample length normalization in its score function (Eq.~\ref{eq:disco_score}).

In contrast, DeepSeek-R1-Distill-Qwen-1.5B, which has been distilled from a reasoning model and already generates long chain-of-thought outputs, exhibits lower within-prompt variance in output lengths. This explains why both Sample Avg and DRPO achieve stable training on this model.

\section{Additional Experimental Results}
\label{app:additional-results}

\subsection{Robustness of the Non-Monotonic Relationship}
\label{app:robustness}

We verify that the non-monotonic relationship is not explained by confounding factors.
The robustness checks in this section are conducted on DeepSeek-R1-Distill-Qwen-1.5B.

\subsubsection{Wall-Clock Time Comparison}
\label{app:walltime}
\begin{figure}[ht]
    \centering
    \includegraphics[width=\textwidth]{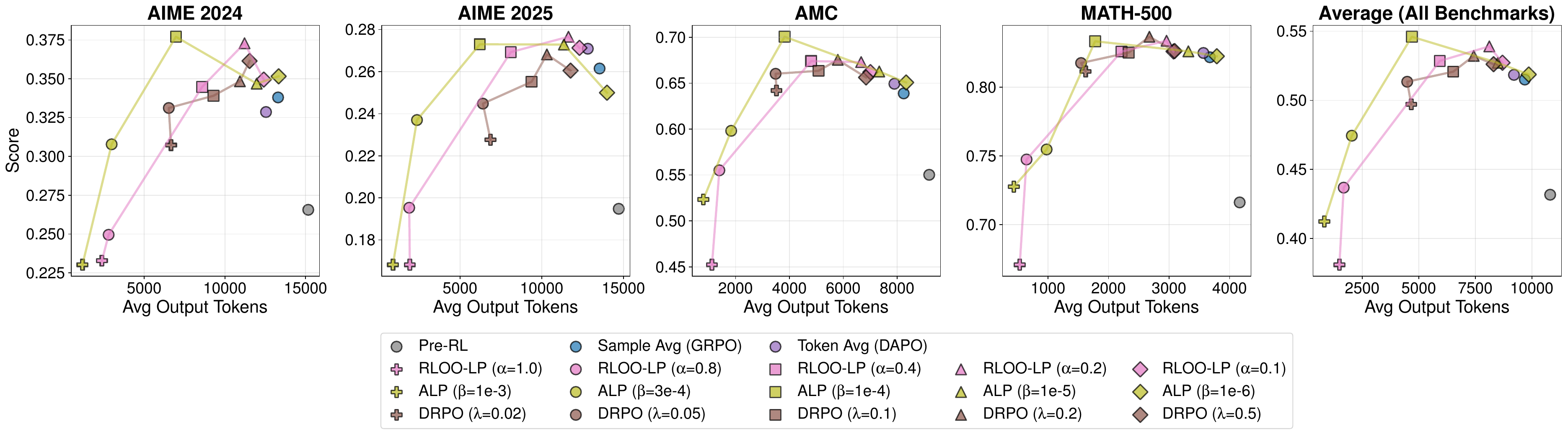}
    \caption{
        \textbf{Score vs.\ average output length for DeepSeek-R1-Distill-Qwen-1.5B at equal wall-clock training time.}
        The non-monotonic relationship persists, indicating that performance degradation with short outputs is not due to reduced training tokens.
    }
    \label{fig:walltime}
\end{figure}

One might hypothesize that shorter outputs lead to lower performance simply because fewer tokens are processed during training, resulting in less learning signal.
To test this, we compare all length-control configurations at our fixed training budget of 576 GPU-hours rather than at equal training steps.
Figure~\ref{fig:walltime} shows that the non-monotonic relationship persists: configurations with strong length penalties still underperform despite having more training steps within the same time budget.
This indicates that the performance degradation in the short-output regime is not an artifact of insufficient training compute.

\subsubsection{Extended Context Length Evaluation}
\label{app:64k}

Another potential explanation for performance degradation at long output lengths is that responses may exceed the evaluation context limit (32K tokens), causing truncation and incorrect answers.
To rule this out, we evaluate with a 64K context length.

\begin{figure}[ht]
    \centering
    \includegraphics[width=\textwidth]{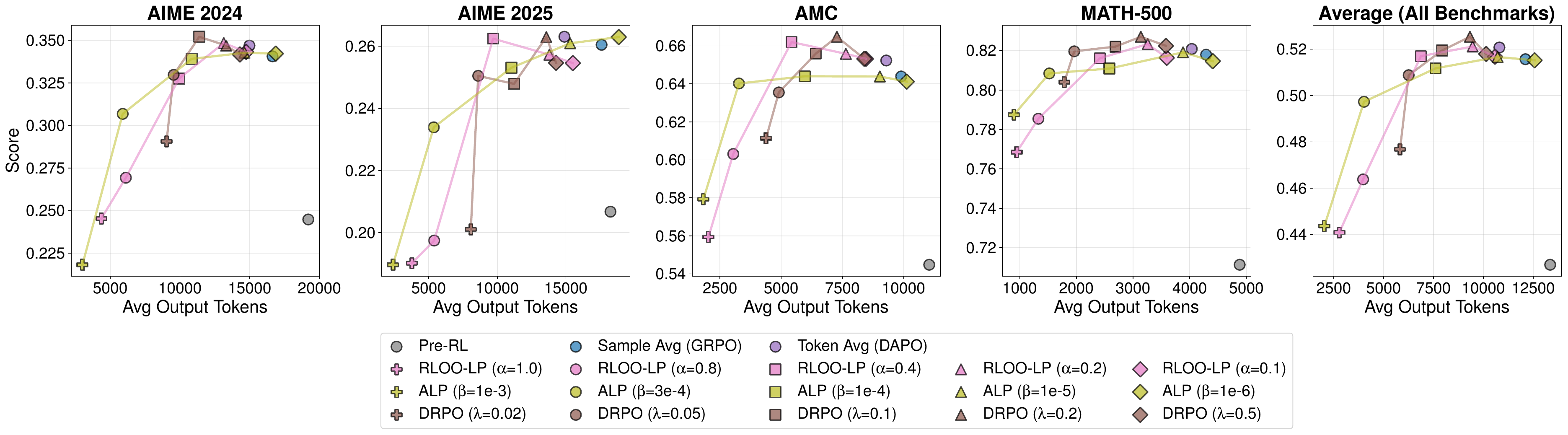}
    \caption{
        \textbf{Score vs.\ average output length for DeepSeek-R1-Distill-Qwen-1.5B with 64K evaluation context length.}
        The non-monotonic pattern is consistent with the 32K evaluation (Figure~\ref{fig:length_performance_math}), confirming that performance degradation at long output lengths is not an artifact of response truncation.
    }
    \label{fig:64k_eval}
\end{figure}

Figure~\ref{fig:64k_eval} shows the score vs.\ output length relationship under 64K evaluation.
The non-monotonic pattern persists, confirming that the performance degradation in the long-output regime is not due to truncation.

\begin{figure}[ht]
    \centering
    \includegraphics[width=\textwidth]{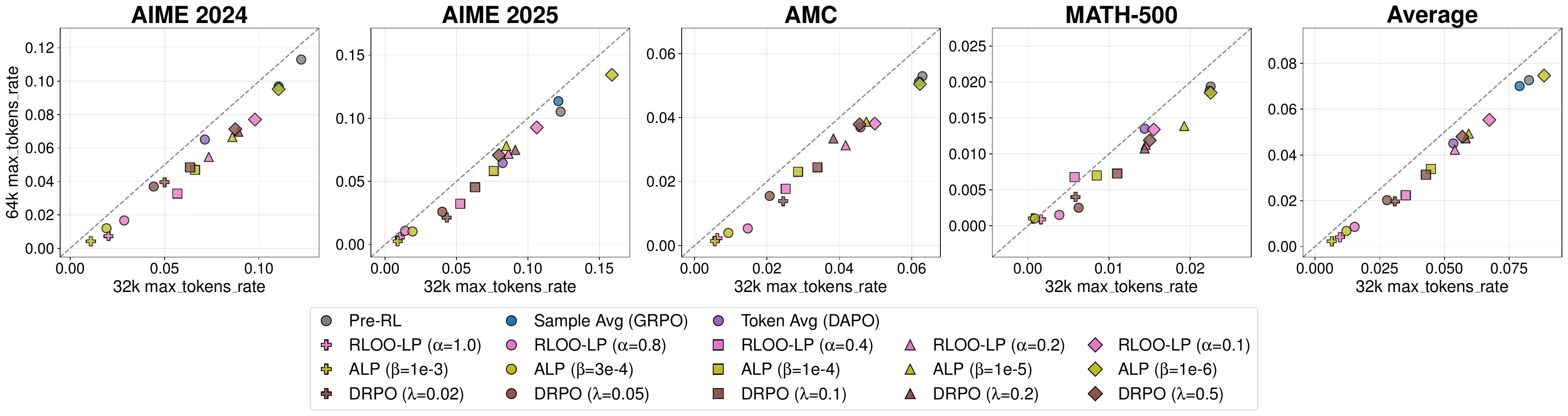}
    \caption{
        \textbf{Truncation rate at 32K vs.\ 64K evaluation context length for DeepSeek-R1-Distill-Qwen-1.5B.}
        Each point represents a method with a specific hyperparameter configuration.
        Points close to the diagonal indicate that extending the context length does not substantially reduce truncation.
    }
    \label{fig:truncation_rate}
\end{figure}

Furthermore, Figure~\ref{fig:truncation_rate} compares the truncation rates between 32K and 64K evaluation contexts across all benchmarks.
The truncation rates remain similar even with the extended context length, indicating that most responses that exceed 32K tokens also exceed 64K tokens.
This suggests that truncation is not the primary cause of performance degradation for long outputs.

These results support our interpretation that the non-monotonic relationship reflects a genuine trade-off in reasoning length, rather than an artifact of the experimental setup.

\subsection{Full Dispersion Analysis}
\label{app:bias-variance-full}

\begin{figure*}[ht]
  \centering
  \includegraphics[width=0.75\linewidth]{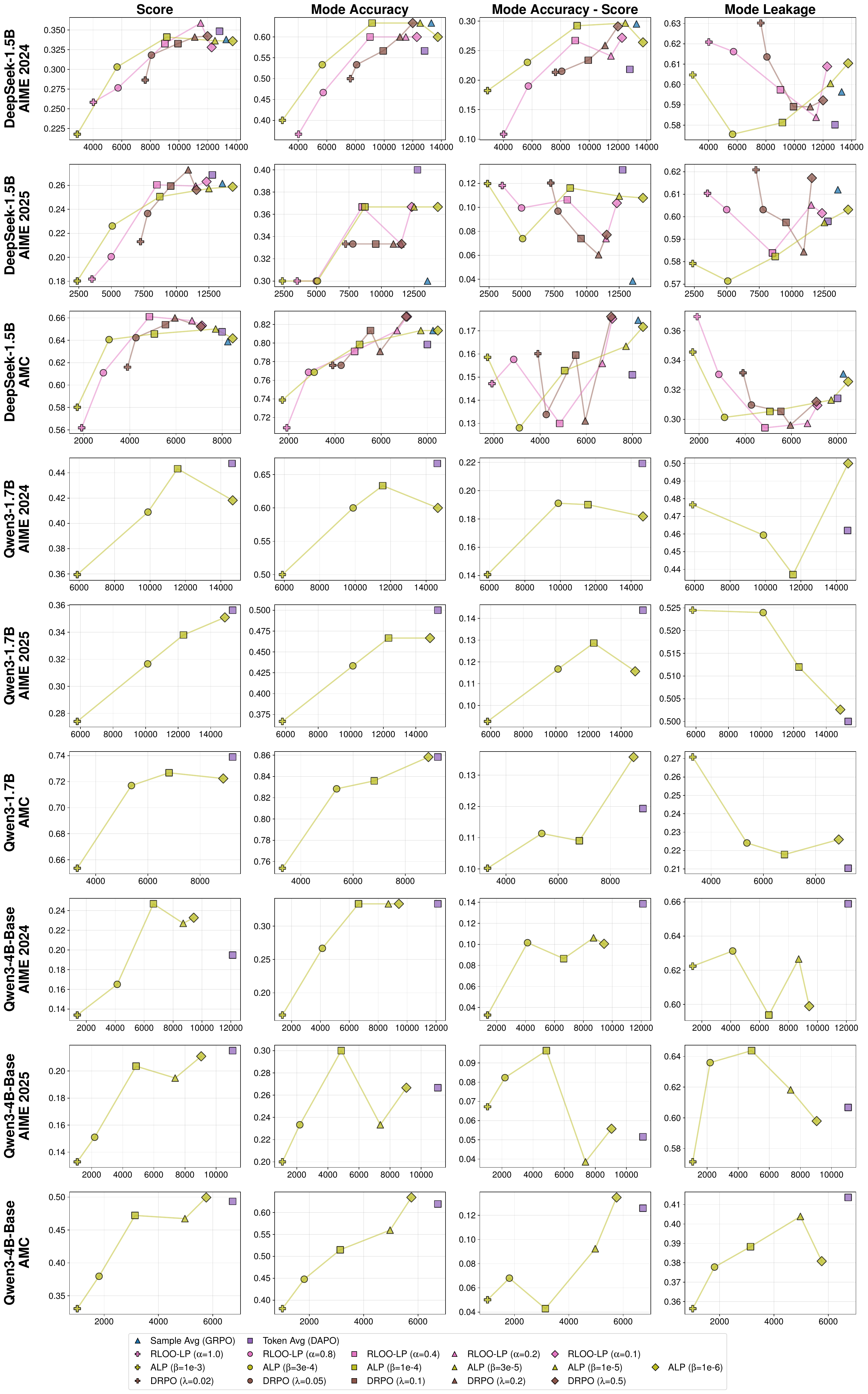}
  \caption{
    \textbf{Score, mode accuracy, mode accuracy minus score, and mode leakage vs.\ average output length on math benchmarks} for the (model, benchmark) pairs not shown in Figure~\ref{fig:ccl}. Across most settings, mode accuracy rises with output length more steadily than score, opening a gap between the two, and mode leakage rises in the long-output regime. The exceptions are AIME 2025 on Qwen3-1.7B and Qwen3-4B-Base, where score itself increases monotonically with length over the range we cover, so the mode--sample divergence has not yet emerged.
  }
  \label{fig:full}
\end{figure*}

Figure~\ref{fig:full} extends the analysis of Section~\ref{sec:explain} to all (model, benchmark) pairs on math not shown in Figure~\ref{fig:ccl}, covering AIME 2024, AIME 2025, and AMC for each of the three models. As in Section~\ref{sec:explain}, we plot score, mode accuracy, the gap between the two, and mode leakage against average output length.

Across most of these settings, the patterns described in Section~\ref{sec:explain} hold. Mode accuracy rises with output length more steadily than score, opening a gap between the two in the long-output regime where score saturates. Mode leakage rises at long lengths, consistent with samples spreading away from an increasingly correct mode, and is U-shaped on many of the settings.

The two AIME 2025 settings on Qwen3-1.7B and Qwen3-4B-Base behave differently. On both models, score increases essentially monotonically with output length over the range we cover (from 0.274 to 0.356 on Qwen3-1.7B, and from 0.133 to 0.215 on Qwen3-4B-Base), and mode accuracy tracks this increase. The mode--sample divergence analyzed in Section~\ref{sec:explain} arises only once score begins to saturate or decline at long lengths; on these two settings that regime has not yet been reached. AIME 2025 is the hardest of our four math benchmarks, so it is plausible that even the longest configurations we trained remain below the length at which dispersion would dominate.

\begin{figure*}[ht]
  \centering
  \includegraphics[width=\linewidth]{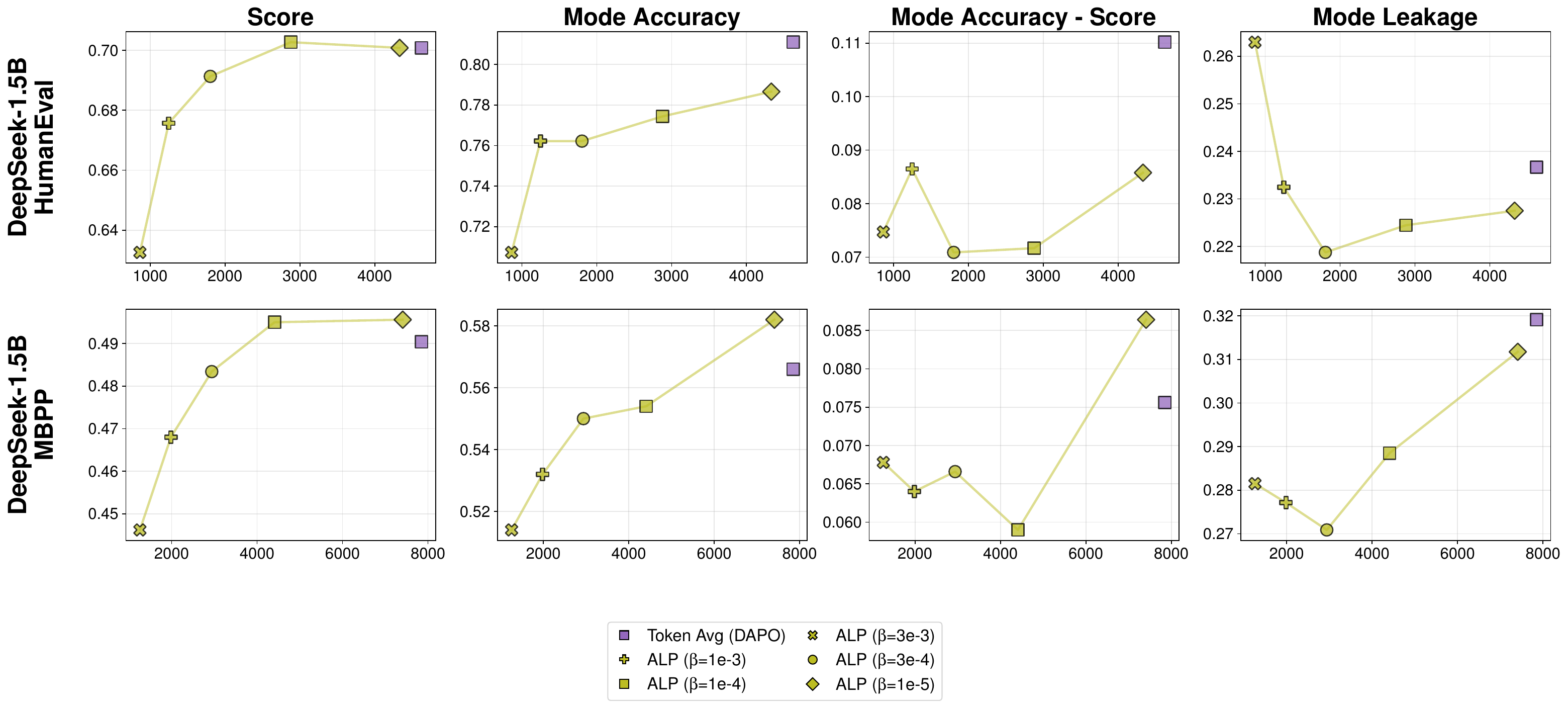}
  \caption{
    \textbf{Score, mode accuracy, mode accuracy minus score, and mode leakage vs.\ average output length on code generation benchmarks} (HumanEval and MBPP) for DeepSeek-R1-Distill-Qwen-1.5B. MBPP shares its underlying problems with MBPP+ used in Figure~\ref{fig:ccl}, and is included here to confirm that the same trend holds. The patterns observed on math also hold on both code benchmarks.
  }
  \label{fig:full-code}
\end{figure*}

\paragraph{Code generation benchmarks.}
Figure~\ref{fig:full-code} shows the same dispersion analysis on HumanEval and MBPP for DeepSeek-R1-Distill-Qwen-1.5B. MBPP+, which we use as the representative code benchmark in the main text, is built from a hand-verified subset of MBPP problems with substantially expanded test cases to catch incorrect solutions passing the original tests. Since the two benchmarks share the same underlying problems, we report MBPP+ in the main text and include MBPP here only to confirm that the same trend holds. The patterns described in Section~\ref{sec:explain} hold on both code benchmarks: mode accuracy rises with output length more steadily than score, and mode leakage rises in the long-output regime.

\end{document}